\documentclass[11pt]{article}

\usepackage[preprint]{acl}

\usepackage{times}
\usepackage{latexsym}

\usepackage[T1]{fontenc}

\usepackage[utf8]{inputenc}

\usepackage{microtype}

\usepackage{inconsolata}

\usepackage{graphicx}

\usepackage{booktabs}
\usepackage{amsmath}
\usepackage{multirow} 

\title{Quantize by Doing: Task-Guided Mixed-Precision Quantization of Vision-Language-Action Models}

\author{
  \textbf{Jiuyi Xu\textsuperscript{1}},
  \textbf{Xiao Hu\textsuperscript{2}},
  \textbf{Meida Chen\textsuperscript{3}},
  \textbf{Song Wang\textsuperscript{4}},
  \textbf{Yang Ye\textsuperscript{2}},
  \textbf{Yangming Shi\textsuperscript{1}}
\\
  \textsuperscript{1}Colorado School of Mines,
  \textsuperscript{2}Northeastern University
\\
  \textsuperscript{3}USC Institute for Creative Technologies,
  \textsuperscript{4}University of Central Florida
}

\begin{document}
\maketitle
\begin{abstract}
Post-training quantization (PTQ) reduces the memory footprint of vision-language-action (VLA) models, but lowering bit widths while maintaining task performance remains challenging.  
Existing work has used sensitivity estimates from calibration data to assign bit widths to different model components.
However, using task success directly to guide mixed-precision allocation is underexplored because closed-loop evaluations are costly and the effect of quantizing one component depends on the precision of other components. 
To this end, we propose \textbf{QbD-VLA}, a task-guided mixed-precision quantization method that uses task success from short closed-loop evaluations to assign weight and activation bit widths while accounting for interactions between model components. 
Experiments covering four VLA models and all four LIBERO suites show strong task performance at average precisions of approximately 4 bits for weights and fewer than 6 bits for activations. On SIMPLER, the gain over uniform quantization is small. On a Franka Research 3 robot arm, simulation-selected allocations show no detected success-rate loss on X-VLA but fail on $\pi_{0.5}$.
These findings support using task success to guide mixed-precision VLA quantization, while highlighting the need for real-world evaluation before deployment.
Code is available at \href{https://github.com/jiuyixu25/QbD-VLA}{\texttt{github.com/jiuyixu25/QbD-VLA}}.

\end{abstract}

\section{Introduction}
\label{sec:intro}

Vision-language-action (VLA) models map camera images and a language instruction to robot actions with backbones of several billion parameters~\citep{kim2024openvla,black2024pi_0,intelligence2025pi_,zheng2025x}.
Post-training quantization (PTQ) is the standard way to fit such a policy onto an on-robot GPU, and the practical question is a mixed-precision allocation: which groups of layers must stay at higher precision while the rest run at lower bit width.
Existing methods choose the allocation from a sensitivity score computed per layer or per group on calibration data, such as Hessian-based sensitivity~\citep{dong2019hawq,deng2023clado}, a reconstruction error ~\citep{nagel2020up,li2021brecq,frantar2022gptq}, or the deviation of predicted actions on fixed inputs ~\citep{xu2026qvla,akbari2026actquant,ranjan2026mixqvla,xu2026daptq}.
Such scores are cheap, but what matters for a robot policy is closed-loop task success, and the two do not agree in general~\citep{vlaquantbench}.
They also score one group at a time, whereas the effect of quantizing a group depends on the precision of the others.
For example, two groups of the $\pi_{0.5}$ action head cost 1.0 and 5.5 success points when quantized alone and 97.5 points together.
Selecting the allocation from task success itself would capture both effects, but closed-loop evaluation is costly, and this problem remains underexplored.

To fill this knowledge gap, we propose \textbf{QbD-VLA}, which selects weight and activation bit widths from closed-loop task success at a bounded cost.
Starting from full precision, it lowers one group by one level at a time. Each candidate is evaluated in simulation with the other groups at their current bit widths, for one to three batches of one episode per task. A sequential test on the Wilson
bounds of the success estimate keeps the move only if the estimate stays above a floor set compared with the full-precision policy.
The allocation ends when no move passes.
On four policies, QbD-VLA finds allocations with nominal weight bit widths of 4.00 to 4.16 and activation bit widths of 4.47 to 5.94.
On $\pi_{0.5}$~\citep{intelligence2025pi_}, OpenVLA-OFT~\citep{kim2025fine}, and X-VLA~\citep{zheng2025x}, uniform W4A4 yields 0\% success, whereas the selected allocations reach 98.5\%, 97.5\%, and 97.0\% against 99.0\%, 100.0\%, and 97.0\% at full precision on LIBERO benchmark~\citep{liu2023libero}. 
At the same time, allocations chosen by action error, activation kurtosis,
or reconstruction error at the same cost reach 27.5\%, 0.0\%, and 52.0\% on $\pi_{0.5}$.
A factorial experiment shows why: lowering the $\pi_{0.5}$ action-expert MLP from A6 to A4 costs 23.5 points when three other groups are at A6 and 0.5 points when they are at A4.
Across all four LIBERO suites, the allocations
stay within 5 points of full precision on average. 
On SIMPLER~\citep{li2024evaluating}, the best allocation reaches 66.8\% against 64.8\% for uniform W4A6.
On a Franka Research 3 robot arm, the frozen allocation succeeds on 13 of 30 layouts for X-VLA against 11 for full precision, but on none of six layouts for $\pi_{0.5}$, where full precision succeeds on all six.
Task success is therefore a practical selection criterion for mixed-precision VLA quantization at a few hundred simulation episodes, while success in simulation does not guarantee execution on hardware. 
Our contributions are the following:
\begin{itemize}
\item We propose QbD-VLA, which selects bit widths for VLA models from closed-loop task success, evaluates each candidate within the current mixed-precision configuration, and bounds the simulation cost with a sequential test.
\item We conduct an evaluation on four VLA policies and all four LIBERO suites against uniform quantization, sensitivity-based allocators, a two-stage allocator, and released VLA quantization methods under the same settings.
\item We analyze why sensitivity-based allocation fails on these policies and test the transfer of simulation-selected allocations to a real robot with two fine-tuned policies.
\end{itemize}

\section{Related Work}
\label{sec:related}

\paragraph{Post-training and mixed-precision quantization.}
Post-training quantization lowers numerical precision
through weight reconstruction or activation-aware scaling
\citep{frantar2022gptq,lin2024awq,xiao2023smoothquant}.
Mixed-precision allocation uses layer sensitivity
\citep{dong2019hawq} or accuracy and hardware feedback
\citep{wang2019haq} to select bit widths.
Additive layer-wise loss approximations omit cross-layer
interactions, which CLADO explicitly models
\citep{deng2023clado}.
All of these score candidates on calibration or validation data fixed in advance.
A control policy instead observes the consequences of its
own earlier actions, so a quantization error can compound
over an episode.
We therefore allocate bit widths from task success measured in closed-loop execution.

\paragraph{Quantization of VLA models.}
VLA quantization methods allocate precision using
action-space sensitivity~\citep{xu2026qvla},
contributions to action prediction~\citep{akbari2026actquant},
internal task evidence~\citep{ranjan2026mixqvla},
or motion-error sensitivity~\citep{xu2026daptq}.
Other approaches address scale calibration
\citep{zhang2026quantvla}, rotation-based W4A4 quantization
\citep{wang2026holoq}, native integer execution
\citep{ho2026foldquantvla}, and precision adjustment using
runtime error feedback~\citep{zheng2026dyq}.
SQAP-VLA combines quantization with token pruning
\citep{fang2025sqap}, while HBVLA and BitVLA investigate
binarization and ternary-weight training, respectively
\citep{yan2026hbvla,wang2025bitvla}.
These methods report closed-loop task performance.
QbD-VLA uses task success to accept candidate group-level
weight and activation allocations, with all other groups
at their current bit widths.

\paragraph{Closed-loop evaluation of quantized policies.}
VLAQuantBench~\citep{vlaquantbench} evaluates quantization effects
in closed loop and shows that the effect of quantizing a group
depends on which other groups are quantized.
We follow its group definitions and develop an allocation method.
A distillation case study documents disagreement between
offline accuracy and execution success~\citep{jia2026anatomy},
while PIVOT-Q~\citep{ruan2026pivotq} uses selective on-policy
distillation to recover post-quantization generalization.
Both train the policy, whereas we only change its bit allocation.
We additionally evaluate whether allocations selected
in simulation transfer to policies fine-tuned for a
real robot.

\begin{figure*}[t]
\centering
\includegraphics[width=\textwidth]{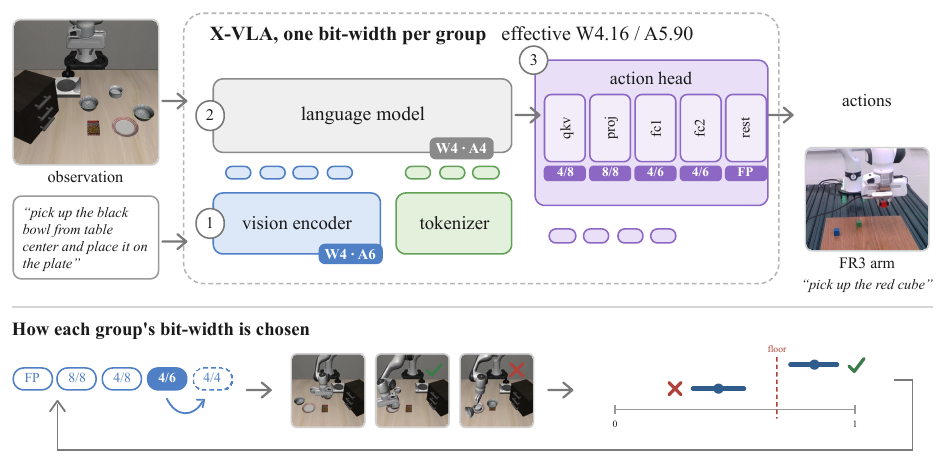}
\caption{The overview of QbD-VLA. A VLA policy (X-VLA) with one weight / activation bit-width per group and the allocation shown has an effective W4.16 / A5.90 and is deployed unchanged on the FR3 arm. A group is lowered
by one level, the policy is rolled out in simulation, and the move is kept only if its success estimate stays above the floor. The search repeats until no move passes.}
\label{fig:overview}
\vspace{-10pt}
\end{figure*}

\section{Method}
\label{sec:method}

QbD-VLA uses closed-loop task success to select the bit widths for quantizing the weights and activations of a pretrained VLA model.
It evaluates each candidate allocation with the other components at their current bit widths, so that bit allocation accounts for interactions between components.
Under a fixed per-candidate budget, evaluation episodes are allocated adaptively according to the uncertainty of each success estimate. Figure~\ref{fig:overview} gives the overview of our method. 

\subsection{Problem Definition}
\label{sec:problem}
We aim to select lower bit widths for a pretrained VLA model while maintaining its task performance.
Following prior VLA quantization work~\citep{xu2026qvla}, we treat the vision encoder, the language backbone, the action head, and any separate vision-language projector as model components.
For bit allocation, we keep these component boundaries and further divide the action head into layer groups following \citet{vlaquantbench}.
Depending on the architecture, the action-head groups cover attention projections, feed-forward layers, and normalization modulators. All quantizable layers within a group share one weight bit width and one activation bit width, so sensitive parts of the action head can keep higher precision without increasing the precision of the entire head.

A configuration $c$ assigns a weight bit width $w_g$ and an activation bit width $a_g$ to each group $g$. The available settings are ordered as
  \[
  \mathrm{FP}
  \rightarrow \mathrm{W8A8}
  \rightarrow \mathrm{W4A8}
  \rightarrow \mathrm{W4A6}
  \rightarrow \mathrm{W4A4},
  \]
where $\mathrm{FP}$ keeps the released precision and $\mathrm{W}w\mathrm{A}a$ denotes $w$-bit weights and $a$-bit linear-input activations. We use fixed round-to-nearest (RTN) quantization~\cite{jacob2018quantization,frantar2022gptq} with no calibration and perform no policy training during
allocation.
Let $N_g$ be the number of quantizable weight parameters in group $g$ and $N=\sum_{g=1}^{G}N_g$. The effective weight and activation bit widths are
\begin{equation}
W(c)=\frac{1}{N}\sum_{g=1}^{G}N_g w_g,\qquad
A(c)=\frac{1}{N}\sum_{g=1}^{G}N_g a_g .
\end{equation}
We define the allocation cost as
\begin{equation}
C(c)=W(c)+A(c).
\label{eq:allocation_cost}
\end{equation}
$A(c)$ is a parameter-weighted bit-width measure, not measured activation memory. Candidates with larger cost reductions are evaluated first, and the same cost is used to select the final configuration.

Let $S(c)$ denote task success under the full evaluation and $c_{\mathrm{FP}}$ the full-precision configuration. Our objective is to reduce $C(c)$ while keeping $S(c)$ close to $S(c_{\mathrm{FP}})$.

\subsection{Task Success Evaluation}
\label{sec:task_evaluation}
Evaluating every candidate under the full evaluation is expensive. We therefore estimate task success from a small number of episodes with a shortened horizon $H$. Each evaluation batch contains one episode per task, and batch $i$ uses initial state $20+i$ of every task, so the allocation episodes are disjoint from the initial states $0$ to $19$ used for final evaluation and every candidate is scored on the same episodes.

For a candidate configuration $c$, let $k_c$ be the number of successful episodes among the $n_c$ evaluated. Its estimated success rate is
\begin{equation}
\widehat{s}_H(c)=\frac{k_c}{n_c}.
\end{equation}
We first evaluate $c_{\mathrm{FP}}$ on all $B$ batches and fix the acceptance threshold
\begin{equation}
f=\tau\,\widehat{s}_H(c_{\mathrm{FP}}),
\label{eq:acceptance_threshold}
\end{equation}
where $\tau$ sets the acceptable decrease compared with the full-precision reference under the same short evaluation. The reference score and the threshold remain fixed throughout allocation. If $\widehat{s}_H(c_{\mathrm{FP}})$ falls below a minimum reference level $s_{\min}$, the short evaluation cannot separate candidates from the threshold and allocation is not started.

Candidates are evaluated one batch at a time. Let $L_z(k,n)$ and $U_z(k,n)$ be the lower and upper Wilson bounds~\citep{wilson1927probable,brown2001interval}, where $z$ is the standard normal quantile that sets their width. We use a smaller quantile $z_{\mathrm{acc}}$ for acceptance than $z_{\mathrm{rej}}$ for rejection. The two errors are not equally costly: a configuration accepted by mistake is still checked by the final evaluation, while a configuration rejected by mistake is dropped from the path and never evaluated again. We therefore require stronger evidence to reject than to accept.
After each batch we accept the candidate if
\begin{equation}
L_{z_{\mathrm{acc}}}(k_c,n_c)\geq f,
\end{equation}
and otherwise reject it if
\begin{equation}
U_{z_{\mathrm{rej}}}(k_c,n_c)\leq f.
\end{equation}
If neither condition holds, we evaluate another batch, up to $B$ batches. If the evaluation has not terminated after $B$ batches, the candidate is accepted if and only if $\widehat{s}_H(c)\geq f$.

\subsection{Task-Guided Bit Allocation}
\label{sec:bit_allocation}

Allocation starts from $c_{\mathrm{FP}}$. At the current configuration $c$, each candidate $c'$ assigns one group to its next lower setting and keeps every other group at its current bit widths. We compute the cost reduction
\begin{equation}
\Delta C(c,c')=C(c)-C(c'),
\end{equation}
evaluate candidates in descending order of $\Delta C$, and accept the first one that satisfies the decision rule of Section~\ref{sec:task_evaluation}. The accepted candidate becomes the current configuration and the candidate set is regenerated.

Each evaluation executes the policy with the candidate allocation applied to all groups, so the decision to lower a group depends on the bit widths already assigned to the other groups. This allows QbD-VLA to account for configuration-dependent effects without assigning each group a fixed sensitivity score.
After a candidate is accepted, every previously rejected group-level move becomes eligible again under the updated allocation. For example, lowering an action-head group to A4 may be rejected at first and accepted after another group has changed.
Evaluation results are cached per configuration and reused whenever the same configuration is evaluated again.
Allocation stops when no candidate at the current configuration is accepted or when the total episode cap is reached. All accepted configurations and their evaluation results are recorded for final selection.

\subsection{Configuration Selection and Evaluation}
\label{sec:configuration_selection}

We select the final configuration from all accepted configurations. The last accepted configuration has the lowest allocation cost, but an earlier configuration may have a higher estimated success rate.
Let $\mathcal{P}$ contain all accepted quantized configurations, excluding the full-precision starting point, and let
\begin{equation}
s_{\max}=\max_{c\in\mathcal{P}}\widehat{s}_H(c).
\end{equation}
We select the lowest-cost configuration whose one-sided Wilson upper bound at $z_{\mathrm{acc}}$ reaches this maximum:
\begin{equation}
c^\star \in
\underset{
c\in\mathcal{P}:\,
U_{z_{\mathrm{acc}}}(k_c,n_c)\geq s_{\max}
}{\arg\min}
C(c).
\label{eq:configuration_selection}
\end{equation}
Equal-cost ties favor the later configuration on the path, and if no quantized configuration is accepted we keep full precision.
Selection uses only the allocation records.
We then freeze $c^\star$ and evaluate it on the independent initial states
with the full task horizon, which gives the reported success rate.

\begin{table*}[t]
\centering
\small
\setlength{\tabcolsep}{5pt}
\begin{tabular}{@{}clrrrrr@{}}
\toprule
Policy & Configuration & Spatial & Object & Goal & Long & Avg. \\
\midrule
\multirow{4}{*}{$\pi_{0.5}$} & Full precision & 99.0 & 99.0 & 94.0 & 95.0 & 96.8 \\
& Uniform W4A4 & 0.0 & 0.0 & 0.0 & 0.0 & 0.0 \\
& Uniform W4A6 & 97.5 & 99.5 & 93.5 & 95.5 & 96.5 \\
& QbD-VLA W4A5.74 & 98.5 & 99.0 & 94.0 & 93.5 & 96.2 \\
\midrule
\multirow{4}{*}{OpenVLA-OFT} & Full precision & 100.0 & 97.5 & 96.0 & 92.5 & 96.5 \\
& Uniform W4A4 & 0.0 & 0.0 & 0.0 & 0.0 & 0.0 \\
& Uniform W4A6 & 23.5 & 2.5 & 1.0 & 62.5 & 22.4 \\
& QbD-VLA W4A5.94 & 97.5 & 94.5 & 94.5 & 80.5 & 91.8 \\
\midrule
\multirow{4}{*}{X-VLA} & Full precision & 97.0 & 98.0 & 97.5 & 98.0 & 97.6 \\
& Uniform W4A4 & 0.0 & 0.0 & 0.0 & 0.0 & 0.0 \\
& Uniform W4A6 & 91.5 & 16.0 & 77.0 & 63.0 & 61.9 \\
& QbD-VLA W4.16A5.90 & 97.0 & 98.0 & 93.0 & 95.0 & 95.8 \\
\midrule
\multirow{4}{*}{$\pi_0$} & Full precision & 78.0 & 87.5 & 84.5 & 48.5 & 74.6 \\
& Uniform W4A4 & 61.0 & 74.0 & 58.0 & 31.0 & 56.0 \\
& Uniform W4A6 & 76.0 & 87.0 & 82.5 & 42.5 & 72.0 \\
& QbD-VLA W4.14A4.47 & 74.5 & 93.0 & 84.0 & 42.5 & 73.5 \\
\bottomrule
\end{tabular}
\caption{Success rate (\%) on the four LIBERO suites. QbD-VLA allocations are selected on LIBERO-Spatial and applied unchanged to the other suites. Their bit widths are parameter-weighted nominal values over all layer groups.}
\label{tab:libero}
\vspace{-10pt}
\end{table*}

\section{Experiments}
\subsection{Experimental Setup}
\label{sec:experimental_setup}

\paragraph{Models and benchmarks.}
We evaluate QbD-VLA on four VLA models:
$\pi_{0.5}$~\citep{intelligence2025pi_}, OpenVLA-OFT~\citep{kim2025fine}, X-VLA~\citep{zheng2025x}, and $\pi_0$~\citep{black2024pi_0}.
We select bit allocations on LIBERO-Spatial and evaluate
the frozen allocations on all four LIBERO suites:
Spatial, Object, Goal, and Long, without further allocation
~\citep{liu2023libero}.
We also evaluate X-VLA on SIMPLER \citep{li2024evaluating},
using the variant-aggregation split for allocation and
the visual-matching split for evaluation.
For real-robot experiments, we apply the allocations selected
in simulation to X-VLA and $\pi_{0.5}$ after fine-tuning
them for a Franka Research 3 (FP3) arm.

\paragraph{Allocation and evaluation.}
Allocations are selected on LIBERO-Spatial with a horizon of 120 steps for $\pi_{0.5}$, OpenVLA-OFT, and $\pi_0$, and 300 steps for X-VLA.
Each batch runs one episode per task, with at most three batches per candidate, Wilson parameters $z_{\mathrm{acc}}=1.28$ and $z_{\mathrm{rej}}=1.96$, a minimum reference score of $s_{\min}=0.5$.
The relative threshold is $\tau=0.75$ for $\pi_{0.5}$ and $0.90$ for the other three models. 
OpenVLA-OFT selects the same allocation under either value.
The same experiment is also run directly on LIBERO-Goal for $\pi_{0.5}$ and OpenVLA-OFT, and on the SIMPLER variant-aggregation split for X-VLA.
Every LIBERO configuration is evaluated on the official protocol with 20 initial states per task.
Full precision (FP), uniform W4A4, uniform W4A6, and the selected allocation are evaluated on all four suites, Spatial, Object, Goal, and Long, with the allocation selected on Spatial applied unchanged to the other three. 
We also compared with released VLA quantization methods (i.e., ActQuant~\cite{akbari2026actquant} and HoloQ-VLA~\cite{wang2026holoq}) under the same evaluation on all four suites with their released code and weights.
SIMPLER allocations are evaluated on the 864 episodes of the visual-matching split, reported by task family.
On the Franka Research 3 (FR3) robot arm, X-VLA and $\pi_{0.5}$ are each evaluated under FP and the QbD-VLA allocation, with 30 trials per condition for X-VLA and six for $\pi_{0.5}$ and six trials per uniform setting (W4A6, W4A8, W8A6 and W8A8) for $\pi_{0.5}$.
We report task success (\%), the nominal weight and activation bits of each allocation, averaged over groups with parameter-count weights and counting unquantized groups as 16 bits. 
The detailed group definition for each model is shown in Appendix~\ref{app:groups}. 
The detailed experimental configuration is shown in Appendix~\ref{app:config}.
Code, evaluation records, and a results video are available at \url{https://github.com/jiuyixu25/QbD-VLA}.

\subsection{Quantization Performance on LIBERO}
\label{sec:libero}

\begin{table}[t]
\centering
\small
\setlength{\tabcolsep}{2pt}
\begin{tabular}{@{}lrrrrrr@{}}
\toprule
Configuration & W / A & Spatial & Object & Goal & Long & Avg. \\
\midrule
FP        & 16 / 16     & 99.0 & 99.0  & 94.0 & 95.0 & 96.8 \\
ActQuant  & 3.09 / 16   & 98.5 & 99.0  & 97.0 & 89.0 & 95.9 \\
HoloQ-VLA & 4 / 4       & 99.0 & 100.0 & 96.5 & 96.5 & 98.0 \\
QbD-VLA   & 4.00 / 5.74 & 98.5 & 99.0  & 94.0 & 93.5 & 96.2 \\
\bottomrule
\end{tabular}
\caption{VLA quantization methods on $\pi_{0.5}$ under our evaluation on the four LIBERO suites, using their released code or weights. The ActQuant checkpoint averages 3.09 weight bits. }
\vspace{-15pt}
\label{tab:released}
\end{table}

\begin{figure}[t]
\centering                                                
\includegraphics[width=\columnwidth]{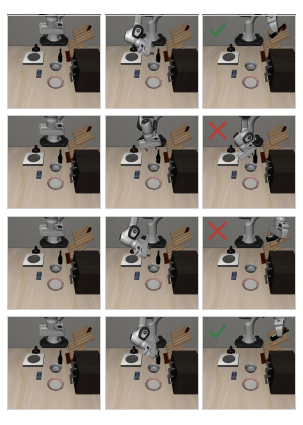}
\caption{$\pi_{0.5}$ rollouts of the LIBERO-Goal task \emph{put the wine bottle on the rack} from the same initial state. Rows, top to bottom: full precision, uniform W4A4, uniform W4A6, and the QbD-VLA allocation. Uniform W4A4 never approaches the bottle and uniform W4A6 drops it short of the rack, whereas the selected allocation completes the task.}
\label{fig:sim_examples}
\vspace{-15pt}
\end{figure}

Table~\ref{tab:libero} evaluates full precision, uniform RTN at W4A4 and W4A6, and the allocation QbD-VLA selects on LIBERO-Spatial on all four suites for every policy.
Uniform W4A4 yields 0\% on every suite for $\pi_{0.5}$, OpenVLA-OFT, and X-VLA, whereas QbD-VLA reaches 98.5\%, 97.5\%, and 97.0\% on Spatial (98.0, 97.5, and 97.0\% on new initial states) with nominal weight bit widths of 4.00 to 4.16 and activation
bit widths of 5.74 to 5.94.
Across all four LIBERO suites, the allocations average
96.2\% on $\pi_{0.5}$, 91.8\% on OpenVLA-OFT, and 95.8\% on X-VLA,
against 96.8, 96.5, and 97.6\% at full precision.
The largest loss is OpenVLA-OFT on Long (80.5\% against 92.5\%).
Allocating directly on LIBERO-Goal recovers the same configurations for $\pi_{0.5}$ and OpenVLA-OFT, so this loss comes from the configuration rather than from the suite it was selected on. Figure~\ref{fig:sim_examples} illustrates some simulation
rollouts.

Whether uniform W4A6 is sufficient depends on the policy.
On $\pi_{0.5}$ it stays within 1.5 points of full precision on every suite (96.5\% average), so there the allocation buys a lower activation budget (5.74 against 6 bits) at matched accuracy.
On OpenVLA-OFT it drops to 23.5, 2.5, and 1.0\% on Spatial, Object, and Goal and to 62.5\% on Long (22.4\% average), whereas the selected configuration, which keeps the action-head output layer at full precision, reaches 97.5, 94.5, 94.5, and 80.5\%.
On X-VLA it reaches 91.5\% on Spatial but 16.0, 77.0, and 63.0\% on Object, Goal, and Long (61.9\% average), against 97.0, 98.0, 93.0, and 95.0\% for the selected allocation.
On $\pi_0$, where W4A4 degrades gradually (31.0 to 74.0\%), the allocation averages 73.5\% against 74.6\% at full precision with 4.14 weight and 4.47 activation bits, below the activation budget of uniform W4A6 (72.0\% average).
The four primary Spatial runs execute 600 to 970 new allocation episodes, and the $\pi_{0.5}$ run additionally reuses 1,070 recorded episodes.

Table~\ref{tab:released} compares QbD-VLA on $\pi_{0.5}$ with two released VLA quantization methods under our evaluation: ActQuant quantizes weights to 3.09 bits on average and retains 16-bit activations, and HoloQ-VLA applies rotation-based
quantization at uniform W4A4.
HoloQ-VLA reaches 99.0\% at uniform W4A4 where RTN scores 0\%, but QbD-VLA studies allocation under a fixed quantizer, and combining it with these quantizers is not evaluated here.

\subsection{Comparison of Bit Allocation Methods}
\label{sec:proxies}

\begin{table}[t]
\centering
\small
\setlength{\tabcolsep}{3pt}
\begin{tabular}{@{}lrrrr@{}}
\toprule
Allocation method & $\pi_{0.5}$ & OpenVLA-OFT & X-VLA & $\pi_0$ \\
\midrule
QbD-VLA               & 98.5 & 97.5 & 97.0 & 74.5 \\
Action error          & 27.5 & 97.5 & 96.0 & 76.0 \\
Activation kurtosis   &  0.0 &  0.0 &  0.0 & 51.0 \\
Reconstruction error  & 52.0 &  0.0 & 91.5 & 60.0 \\
\midrule
UFR                   & 99.0 & 99.0 & 87.0 & 68.0 \\
\bottomrule
\end{tabular}
\caption{Success rate (\%) on LIBERO-Spatial. The three sensitivity-based rows share a greedy allocator and the target cost of the QbD-VLA allocation of each policy. UFR reaches its own cost: W4.00 / A5.90, W8.00 / A8.00, W4.00 / A5.62, and W4.00 / A4.54. }
\label{tab:allocation}
\vspace{-10pt}
\end{table}

\begin{table*}[t]
\centering
\small
\setlength{\tabcolsep}{3pt}
\begin{tabular}{@{}lrrrrrr@{}}
\toprule
X-VLA configuration & W / A & Coke can & Move near & Drawer & Put in drawer & Overall \\
\midrule
FP & 16 / 16 & 90.7 & 84.2 & 65.7 & 29.6 & 75.0 \\
Uniform W4A8 & 4 / 8 & 94.0 & 83.8 & 64.4 & 28.7 & 75.6 \\
Uniform W4A6 & 4 / 6 & 87.0 & 75.4 & 52.3 & 4.6 & 64.8 \\
Uniform W4A4 & 4 / 4 & 0.0 & 1.7 & 13.9 & 0.0 & 3.9 \\
QbD-VLA, 24 tasks, $\tau=0.90$ & 4.03 / 5.33 & 87.7 & 65.8 & 44.4 & 0.0 & 59.8 \\
QbD-VLA, 40 tasks, $\tau=0.95$ & 4.00 / 5.88 & 89.3 & 84.2 & 43.5 & 12.0 & 66.8 \\
\bottomrule
\end{tabular}
\caption{X-VLA on the SIMPLER visual-matching split. }
\label{tab:simpler}
\vspace{-5pt}
\end{table*}

Table~\ref{tab:allocation} compares QbD-VLA with allocations chosen by three sensitivity-based allocation baselines inspired by prior work: open-loop action error ~\citep{xu2026qvla,akbari2026actquant,xu2026daptq}, activation kurtosis~\citep{bondarenko2023quantizable,akhondzadeh2025kurtail,zhang2025layersensitive}, and weight reconstruction error \citep{nagel2020up,li2021brecq,frantar2022gptq}.
They use the same layer groups and quantization strategies as QbD-VLA.
We also compare with a two-stage allocator, uniform allocation followed by group relaxation (UFR). 
It first evaluates uniform configurations in decreasing precision, W8A8, W4A8, W4A6, and W4A4, with the same closed-loop acceptance test as QbD-VLA, and keeps the lowest-precision uniform configuration that passes.
It then evaluates the bit width of one group at a time by one level, re-evaluates each reduced configuration, and finally verifies all accepted single-group reductions applied together.
UFR follows the same evaluation as QbD-VLA but is not constrained to the same cost.
On $\pi_{0.5}$ the sensitivity-based allocations reach 27.5, 0.0, and 52.0\% against 98.5\% for QbD-VLA, and on OpenVLA-OFT two of the three reach 0\%.
On X-VLA and $\pi_0$ the action-error allocation reaches 96.0\% and 76.0\%, against 97.0\% and 74.5\% for QbD-VLA.
UFR achieves similar success to QbD-VLA on $\pi_{0.5}$ at a higher cost, reaches 99.0\% on OpenVLA-OFT at about W8A8, and reaches 87.0\% on X-VLA and 68.0\% on $\pi_0$.
A factorial experiment on $\pi_{0.5}$ shows why a sensitivity measured once per group can mislead.
With the attention, adaRMS-post, and I/O groups at A6, changing the MLP group from A6 to A4 changes success from 97.5\% to 74.0\%.
With those groups at A4, the same MLP change gives 99.0\% and 98.5\%, respectively.
(The example in Section~\ref{sec:intro} is from the same policy: with every other group at full precision, W4A4 on the action-head attention projections alone gives 98.0\%, on the modulators alone 93.5\%, and on both together 1.5\%, against 99.0\% at full precision.)
These results show that the effect of a group's bit width depends on the precision of other groups, supporting evaluation within the current mixed-precision configuration.

\subsection{Robustness of the Allocation}
\label{sec:reliability}

The allocation accepts or rejects a configuration after at most three short batches, so we re-evaluated all 80 configurations accepted across the four LIBERO-Spatial runs (25, 18, 20, and 17 for $\pi_{0.5}$, OpenVLA-OFT, X-VLA, and $\pi_0$) with 100
full-horizon episodes each.
The lowest success along each accepted path is 94\% for $\pi_{0.5}$, 98\% for OpenVLA-OFT, 88\% for X-VLA, and 65\% for $\pi_0$, and the path means are 97.8, 99.1, 96.5, and 73.5\%, against full-precision references of 99, 100, 98, and 74\% on the same
initial states. 
Hence, the short batches never admitted a configuration more than 10 points below full precision, even though the short-horizon estimates correlate weakly with full-horizon success.
The endpoint is also stable to the threshold: across $\tau \in \{0.50, 0.75, 0.90, 0.95\}$ on $\pi_{0.5}$, the selected configurations lie between 4.00 / 5.74 and 4.10 / 6.32 (W / A) with 98.0 to 98.5\% success, where a higher threshold stops earlier
(14 accepted moves at $\tau=0.95$ against 25 at $\tau=0.75$) at a higher activation cost.
A single-batch variant selects the same configuration as the reference run with 760 instead of 970 allocation episodes.

On SIMPLER, allocation is performed on the variant-aggregation split, which varies lighting, backgrounds, and distractors, and Table~\ref{tab:simpler} reports the disjoint visual-matching split, which replicates the real scenes, by task family.
Uniform W4A8 matches full precision on every family (75.6\% overall against 75.0\%), uniform W4A6 loses 13 points on the drawer tasks and 25 on put in drawer, and uniform W4A4 collapses.
The allocation selected on 24 tasks with $\tau=0.90$ (A5.33) reaches 59.8\%, keeping pick coke can within 3 points of full precision but falling below uniform W4A6 on the other three families. 
Selecting on all 40 tasks with $\tau=0.95$ gives A5.88 and
66.8\%, matching full precision on pick coke can and move near but still 22 points below it on the drawer tasks and 18 on put in drawer, which full precision itself completes only 29.6\% of the time.
The two allocations trade families rather than recover all of them, and the gain over uniform W4A6 is 2.0 points at a similar activation cost, because the search observes only the variant-aggregation split while the drawer failures appear only under
visual matching.

\begin{table}[t]
\centering
\small
\setlength{\tabcolsep}{4pt}
\begin{tabular}{@{}clrr@{}}
\toprule
Model & Configuration & Success & Close cmd. \\
\midrule
\multirow{2}{*}{X-VLA}
& FP & 11/30 & 28/30 \\
& QbD-VLA        & 13/30 & 29/30 \\
\midrule
\multirow{2}{*}{$\pi_{0.5}$}
& FP & 6/6 & 6/6 \\
& QbD-VLA        & 0/6 & 0/6 \\
\bottomrule
\end{tabular}
\caption{
Transfer of frozen allocations to the FR3 arm.
}
\label{tab:robot}
\vspace{-15pt}
\end{table}

\begin{table}[t]
\centering
\small
\setlength{\tabcolsep}{8pt}
\begin{tabular}{@{}lrr@{}}
\toprule
Configuration & Success & Close cmd. \\
\midrule
Uniform W4A6 & 0/6 & 0/6 \\
Uniform W4A8 & 1/6 & 1/6 \\
Uniform W8A6 & 0/6 & 0/6 \\
Uniform W8A8 & 5/6 & 6/6 \\
\bottomrule
\end{tabular}
\caption{
Uniform-precision diagnostics for $\pi_{0.5}$ on the FR3 arm.
}
\vspace{-35pt}
\label{tab:robot_uniform}
\end{table}

\begin{figure*}[t]
\centering
\includegraphics[width=\textwidth]{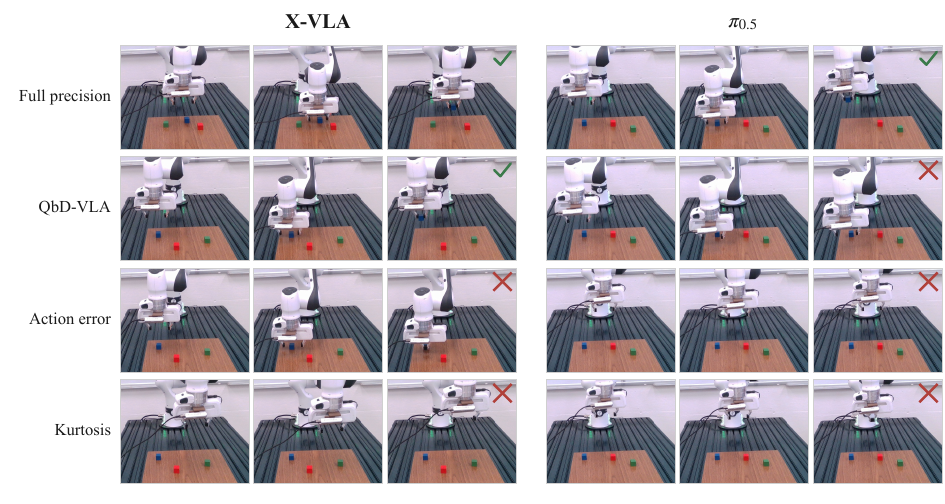}
\caption{Some example trials on the FR3 arm.}
\label{fig:robot_examples}
\vspace{-5pt}
\end{figure*}

\subsection{Transfer to a Real Robot}
\label{sec:robot}

We evaluate whether an allocation selected in simulation
remains effective on the real robot.
We fine-tune X-VLA and $\pi_{0.5}$ with LoRA on the same
411 demonstrations of FR3 arm picking up the cube of different colors.
Our primary comparison is between full precision and the
allocation selected by QbD-VLA on LIBERO-Spatial.
Table~\ref{tab:robot} reports the transfer results.
On X-VLA, QbD-VLA succeeds on 13 of 30 trials, compared
with 11 for full precision.
On $\pi_{0.5}$, full precision succeeds on all six trials
and QbD-VLA on none.
QbD-VLA never commands gripper closure, with every trial's
peak gripper output below the 0.5 threshold.
The action-error and kurtosis allocations obtained under
the same target budget as QbD-VLA also fail on all six
trials.
Transfer failure is therefore observed for all three
evaluated low-bit allocations.
Table~\ref{tab:robot_uniform} reports subsequent
uniform-precision diagnostics on $\pi_{0.5}$.
With the same six trials and control settings, uniform
W8A8 succeeds on five layouts, compared with one at W4A8
and none at W4A6 or W8A6.
W8A8 has an allocation cost $W(c)+A(c)$ of 16, compared
with 9.74 for the transferred QbD-VLA allocation.
Uniform W4A6 fails on the robot despite reaching
97.5\% on LIBERO-Spatial, where QbD-VLA reaches 98.5\%.
These diagnostics show recovery at a higher bit budget
among the evaluated configurations.
Figure~\ref{fig:robot_examples} shows example trials.
The transfer results differ across the two fine-tuned
policies.
Strong performance in simulation does not guarantee
successful execution on real hardware, so the selected
allocation must also be evaluated on the target real robot
and task.

\section{Conclusion}

We presented QbD-VLA, a task-guided mixed-precision quantization method for vision-language-action models.
It selects bit widths from closed-loop task success without training or calibration, scoring each candidate with the other groups at their current precision to capture interactions between groups.
On four VLA policies, it finds allocations with about 4-bit weights and under 6-bit activations.
Where uniform W4A4 yields 0\%, the selected allocations reach 97.0 to 98.5\% on LIBERO-Spatial and transfer to the other suites with comparable averages, whereas sensitivity-based allocations at the same cost fail on at least one policy each and
uniform W4A6 collapses on OpenVLA-OFT.
Gains on SIMPLER are small because its failures appear only on the evaluation split.
On a real FR3 robot the frozen allocation holds on X-VLA but fails on $\pi_{0.5}$ despite near-full-precision success in simulation.
Task success is therefore a practical allocation criterion, but the selected configuration must still be evaluated on the target robot and task.

\section*{Limitations}

QbD-VLA considers predefined layer groups and candidate bit widths under round-to-nearest quantization without calibration.
These choices define the allocation space studied here, and alternative groupings or quantizers may lead to different configurations. 
Joint optimization of grouping, quantizer, and bit allocation remains for future work.
The greedy procedure evaluates candidates along the current allocation path and does not guarantee global optimality.
Allocation requires a closed-loop simulator and a task suite for the target policy, and costs 600 to 970 simulation episodes per run, so it does not apply where no simulator exists.
Each reported evaluation uses 200 episodes, so differences of one to two points lie within sampling noise, and the sequential test controls each candidate decision but gives no simultaneous guarantee across the decisions of a run.
Reported bit costs characterize nominal precision assignments. 
Actual memory use and integer-kernel latency depend on the deployment implementation and require separate measurement.
Robot experiments assess the transfer of frozen allocations to two fine-tuned policies on one cube-picking task with 30 and 6 layouts, and are exploratory.
They provide evidence about configuration transfer, but allocation directly on hardware remains untested, and the mechanism behind the $\pi_{0.5}$ failure is not identified.
The uniform-precision diagnostics examine a limited set of settings and do not establish the minimum bit budget for deployment.

\bibliography{custom}

\begin{thebibliography}{35}
\providecommand{\natexlab}[1]{#1}

\bibitem[{Akbari et~al.(2026)Akbari, Akbari, Eskandar, Tan, Chen, Luo,
  Pangaribuan, Zhang, Dy, Yuan, Lin, Liu, Ioannidis, and
  Wang}]{akbari2026actquant}
Arash Akbari, Arman Akbari, Masih Eskandar, Qitao Tan, Yixiao Chen, Jingwu Luo,
  Bertha Pangaribuan, Liyun Zhang, Jennifer Dy, Geng Yuan, Xue Lin, Gaowen Liu,
  Stratis Ioannidis, and Yanzhi Wang. 2026.
\newblock \href {https://arxiv.org/abs/2605.24011v3} {{ActQuant}: Sub-4-bit
  action-guided quantization for vision-language-action models}.
\newblock \emph{arXiv preprint arXiv:2605.24011}.

\bibitem[{Akhondzadeh et~al.(2025)Akhondzadeh, Bojchevski, Eleftheriou, and
  Dazzi}]{akhondzadeh2025kurtail}
Mohammad~Sadegh Akhondzadeh, Aleksandar Bojchevski, Evangelos Eleftheriou, and
  Martino Dazzi. 2025.
\newblock \href {https://doi.org/10.18653/v1/2025.findings-emnlp.943}
  {{KurTail}: Kurtosis-based {LLM} quantization}.
\newblock In \emph{Findings of the Association for Computational Linguistics:
  EMNLP 2025}, pages 17404--17419. Association for Computational Linguistics.

\bibitem[{Black et~al.(2024)Black, Brown, Driess et~al.}]{black2024pi_0}
Kevin Black, Noah Brown, Danny Driess, et~al. 2024.
\newblock \href {https://arxiv.org/abs/2410.24164} {{$\pi_0$}: A
  vision-language-action flow model for general robot control}.
\newblock \emph{arXiv preprint arXiv:2410.24164}.

\bibitem[{Bondarenko et~al.(2023)Bondarenko, Nagel, and
  Blankevoort}]{bondarenko2023quantizable}
Yelysei Bondarenko, Markus Nagel, and Tijmen Blankevoort. 2023.
\newblock Quantizable transformers: Removing outliers by helping attention
  heads do nothing.
\newblock In \emph{Advances in Neural Information Processing Systems},
  volume~36, pages 75067--75096.

\bibitem[{Brown et~al.(2001)Brown, Cai, and DasGupta}]{brown2001interval}
Lawrence~D. Brown, T.~Tony Cai, and Anirban DasGupta. 2001.
\newblock Interval estimation for a binomial proportion.
\newblock \emph{Statistical Science}, 16(2):101--133.

\bibitem[{Deng et~al.(2025)Deng, Sharify, Wang, and Orshansky}]{deng2023clado}
Zihao Deng, Sayeh Sharify, Xin Wang, and Michael Orshansky. 2025.
\newblock \href {https://doi.org/10.1109/DAC63849.2025.11132777}
  {Mixed-precision quantization for deep vision models with integer quadratic
  programming}.
\newblock In \emph{Proceedings of the 62nd Annual Design Automation Conference
  ({DAC})}.

\bibitem[{Dong et~al.(2019)Dong, Yao, Gholami, Mahoney, and
  Keutzer}]{dong2019hawq}
Zhen Dong, Zhewei Yao, Amir Gholami, Michael~W. Mahoney, and Kurt Keutzer.
  2019.
\newblock \href {https://arxiv.org/abs/1905.03696} {{HAWQ}: {Hessian AWare}
  quantization of neural networks with mixed-precision}.
\newblock In \emph{Proceedings of the IEEE/CVF International Conference on
  Computer Vision}, pages 293--302.

\bibitem[{Fang et~al.(2025)Fang, Liu, Du, Yang, and Du}]{fang2025sqap}
Hengyu Fang, Yijiang Liu, Yuan Du, Huanrui Yang, and Li~Du. 2025.
\newblock \href {https://arxiv.org/abs/2509.09090} {{SQAP-VLA}: A synergistic
  quantization-aware pruning framework for high-performance
  vision-language-action models}.
\newblock \emph{arXiv preprint arXiv:2509.09090}.

\bibitem[{Frantar et~al.(2022)Frantar, Ashkboos, Hoefler, and
  Alistarh}]{frantar2022gptq}
Elias Frantar, Saleh Ashkboos, Torsten Hoefler, and Dan Alistarh. 2022.
\newblock \href {https://arxiv.org/abs/2210.17323} {{GPTQ}: Accurate
  post-training quantization for generative pre-trained transformers}.
\newblock \emph{arXiv preprint arXiv:2210.17323}.

\bibitem[{Ho et~al.(2026)Ho, Nguyen, Nguyen, Duong, Le, Guo, Ngo, and
  Le}]{ho2026foldquantvla}
Hung~T. Ho, Khanh~D. Nguyen, Quang~D. Nguyen, Thanh~Q. Duong, Ngan Le, Meng
  Guo, Vien~A. Ngo, and An~T. Le. 2026.
\newblock \href {https://arxiv.org/abs/2609.24433} {{FoldQuantVLA}: Native
  low-bit quantization of vision-language-action models via consistent
  folding}.
\newblock \emph{arXiv preprint arXiv:2609.24433}.

\bibitem[{Jacob et~al.(2018)Jacob, Kligys, Chen, Zhu, Tang, Howard, Adam, and
  Kalenichenko}]{jacob2018quantization}
Benoit Jacob, Skirmantas Kligys, Bo~Chen, Menglong Zhu, Matthew Tang, Andrew
  Howard, Hartwig Adam, and Dmitry Kalenichenko. 2018.
\newblock Quantization and training of neural networks for efficient
  integer-arithmetic-only inference.
\newblock In \emph{Proceedings of the IEEE Conference on Computer Vision and
  Pattern Recognition}, pages 2704--2713.

\bibitem[{Jia(2026)}]{jia2026anatomy}
Fengze Jia. 2026.
\newblock \href {https://arxiv.org/abs/2609.23048} {Anatomy of a closed-loop
  collapse: A causal case study of a compressed {VLA} policy}.
\newblock \emph{arXiv preprint arXiv:2609.23048}.

\bibitem[{Kim et~al.(2025)Kim, Finn, and Liang}]{kim2025fine}
Moo~Jin Kim, Chelsea Finn, and Percy Liang. 2025.
\newblock \href {https://arxiv.org/abs/2502.19645} {Fine-tuning
  vision-language-action models: Optimizing speed and success}.
\newblock \emph{arXiv preprint arXiv:2502.19645}.

\bibitem[{Kim et~al.(2024)Kim, Pertsch, Karamcheti, Xiao, Balakrishna, Nair,
  Rafailov, Foster, Lam, Sanketi et~al.}]{kim2024openvla}
Moo~Jin Kim, Karl Pertsch, Siddharth Karamcheti, Ted Xiao, Ashwin Balakrishna,
  Suraj Nair, Rafael Rafailov, Ethan Foster, Grace Lam, Pannag Sanketi, et~al.
  2024.
\newblock \href {https://arxiv.org/abs/2406.09246} {{OpenVLA}: An open-source
  vision-language-action model}.
\newblock \emph{arXiv preprint arXiv:2406.09246}.

\bibitem[{Li et~al.(2024)Li, Hsu, Gu, Pertsch, Mees, Walke, Fu, Lunawat, Sieh,
  Kirmani et~al.}]{li2024evaluating}
Xuanlin Li, Kyle Hsu, Jiayuan Gu, Karl Pertsch, Oier Mees, Homer~Rich Walke,
  Chuyuan Fu, Ishikaa Lunawat, Isabel Sieh, Sean Kirmani, et~al. 2024.
\newblock \href {https://arxiv.org/abs/2405.05941} {Evaluating real-world robot
  manipulation policies in simulation}.
\newblock \emph{arXiv preprint arXiv:2405.05941}.

\bibitem[{Li et~al.(2021)Li, Gong, Tan, Yang, Hu, Zhang, Yu, Wang, and
  Gu}]{li2021brecq}
Yuhang Li, Ruihao Gong, Xu~Tan, Yang Yang, Peng Hu, Qi~Zhang, Fengwei Yu, Wei
  Wang, and Shi Gu. 2021.
\newblock \href {https://arxiv.org/abs/2102.05426} {{BRECQ}: Pushing the limit
  of post-training quantization by block reconstruction}.
\newblock \emph{arXiv preprint arXiv:2102.05426}.

\bibitem[{Lin et~al.(2024)Lin, Tang, Tang, Yang, Chen, Wang, Xiao, Dang, Gan,
  and Han}]{lin2024awq}
Ji~Lin, Jiaming Tang, Haotian Tang, Shang Yang, Wei-Ming Chen, Wei-Chen Wang,
  Guangxuan Xiao, Xingyu Dang, Chuang Gan, and Song Han. 2024.
\newblock \href {https://arxiv.org/abs/2306.00978} {{AWQ}: Activation-aware
  weight quantization for on-device {LLM} compression and acceleration}.
\newblock In \emph{Proceedings of Machine Learning and Systems}, volume~6,
  pages 87--100.

\bibitem[{Liu et~al.(2023)Liu, Zhu, Gao, Feng, Liu, Zhu, and
  Stone}]{liu2023libero}
Bo~Liu, Yifeng Zhu, Chongkai Gao, Yihao Feng, Qiang Liu, Yuke Zhu, and Peter
  Stone. 2023.
\newblock {LIBERO}: Benchmarking knowledge transfer for lifelong robot
  learning.
\newblock In \emph{Advances in Neural Information Processing Systems},
  volume~36, pages 44776--44791.

\bibitem[{Nagel et~al.(2020)Nagel, Amjad, Van~Baalen, Louizos, and
  Blankevoort}]{nagel2020up}
Markus Nagel, Rana~Ali Amjad, Mart Van~Baalen, Christos Louizos, and Tijmen
  Blankevoort. 2020.
\newblock \href {https://proceedings.mlr.press/v119/nagel20a.html} {Up or down?
  {Adaptive} rounding for post-training quantization}.
\newblock In \emph{Proceedings of the 37th International Conference on Machine
  Learning}, volume 119 of \emph{Proceedings of Machine Learning Research},
  pages 7197--7206. PMLR.

\bibitem[{{Physical Intelligence} et~al.(2025){Physical Intelligence}, Black,
  Brown, Darpinian, Dhabalia, Driess, Esmail, Equi, Finn, Fusai
  et~al.}]{intelligence2025pi_}
{Physical Intelligence}, Kevin Black, Noah Brown, James Darpinian, Karan
  Dhabalia, Danny Driess, Adnan Esmail, Michael Equi, Chelsea Finn, Niccolo
  Fusai, et~al. 2025.
\newblock \href {https://arxiv.org/abs/2504.16054} {{$\pi_{0.5}$}: A
  vision-language-action model with open-world generalization}.
\newblock \emph{arXiv preprint arXiv:2504.16054}.

\bibitem[{Ranjan and Savakis(2026)}]{ranjan2026mixqvla}
Navin Ranjan and Andreas Savakis. 2026.
\newblock \href {https://arxiv.org/abs/2606.19565} {{Mix-QVLA}:
  Task-evidence-aware mixed-precision quantization of vision-language-action
  models}.
\newblock \emph{arXiv preprint arXiv:2606.19565}.

\bibitem[{Ruan et~al.(2026)Ruan, Gao, Wang, Liu, {Zhoxizhuoma}, Ren, and
  Zheng}]{ruan2026pivotq}
Shen Ruan, Wenchang Gao, Jin Wang, Siao Liu, {Zhoxizhuoma}, Dongchun Ren, and
  Xin Zheng. 2026.
\newblock \href {https://arxiv.org/abs/2610.05745} {Beyond in-distribution
  preservation: Recovering generalization in quantized {VLAs} via
  vulnerability-oriented tuning}.
\newblock \emph{arXiv preprint arXiv:2610.05745}.

\bibitem[{Wang et~al.(2025)Wang, Xiong, Wang, and Chen}]{wang2025bitvla}
Hongyu Wang, Chuyan Xiong, Ruiping Wang, and Xilin Chen. 2025.
\newblock \href {https://arxiv.org/abs/2506.07530} {{BitVLA}: 1-bit
  vision-language-action models for robotics manipulation}.
\newblock \emph{arXiv preprint arXiv:2506.07530}.

\bibitem[{Wang et~al.(2019)Wang, Liu, Lin, Lin, and Han}]{wang2019haq}
Kuan Wang, Zhijian Liu, Yujun Lin, Ji~Lin, and Song Han. 2019.
\newblock \href {https://arxiv.org/abs/1811.08886} {{HAQ}: Hardware-aware
  automated quantization with mixed precision}.
\newblock In \emph{Proceedings of the IEEE/CVF Conference on Computer Vision
  and Pattern Recognition}, pages 8612--8620.

\bibitem[{Wang et~al.(2026)Wang, Li, Lyu, Liu, Yang, Zhao, Cui, Chang, and
  Lu}]{wang2026holoq}
Xinyu Wang, Mingze Li, Sicheng Lyu, Dongxiu Liu, Kaicheng Yang, Ziyu Zhao,
  Yufei Cui, Xiao-Wen Chang, and Peng Lu. 2026.
\newblock \href {https://arxiv.org/abs/2605.28803v3} {{HoloQ-VLA}: Uniform
  {W4A4} quantization of vision-language-action models}.
\newblock \emph{arXiv preprint arXiv:2605.28803}.

\bibitem[{Wilson(1927)}]{wilson1927probable}
Edwin~B. Wilson. 1927.
\newblock Probable inference, the law of succession, and statistical inference.
\newblock \emph{Journal of the American Statistical Association},
  22(158):209--212.

\bibitem[{Xiao et~al.(2023)Xiao, Lin, Seznec, Wu, Demouth, and
  Han}]{xiao2023smoothquant}
Guangxuan Xiao, Ji~Lin, Mickael Seznec, Hao Wu, Julien Demouth, and Song Han.
  2023.
\newblock \href {https://proceedings.mlr.press/v202/xiao23c.html}
  {{SmoothQuant}: Accurate and efficient post-training quantization for large
  language models}.
\newblock In \emph{Proceedings of the 40th International Conference on Machine
  Learning}, volume 202 of \emph{Proceedings of Machine Learning Research},
  pages 38087--38099. PMLR.

\bibitem[{Xu et~al.(2026{\natexlab{a}})Xu, Jin, Chen, Wang, Sui, and
  Shi}]{vlaquantbench}
Jiuyi Xu, Qing Jin, Meida Chen, Song Wang, Yang Sui, and Yangming Shi.
  2026{\natexlab{a}}.
\newblock \href {https://arxiv.org/abs/2609.25376} {{VLAQuantBench}:
  Closed-loop evaluation of post-training quantization for
  vision-language-action models}.
\newblock \emph{arXiv preprint arXiv:2609.25376}.

\bibitem[{Xu et~al.(2026{\natexlab{b}})Xu, Wang, Li, Zhu, and
  Shen}]{xu2026daptq}
Siyuan Xu, Tianshi Wang, Fengling Li, Lei Zhu, and Heng~Tao Shen.
  2026{\natexlab{b}}.
\newblock \href {https://arxiv.org/abs/2604.11572} {{DA-PTQ}: Drift-aware
  post-training quantization for efficient vision-language-action models}.
\newblock \emph{arXiv preprint arXiv:2604.11572}.

\bibitem[{Xu et~al.(2026{\natexlab{c}})Xu, Yang, Fan, Liu, Li, Li, and
  Zhang}]{xu2026qvla}
Yuhao Xu, Yantai Yang, Zhenyang Fan, Yufan Liu, Yuming Li, Bing Li, and Zhipeng
  Zhang. 2026{\natexlab{c}}.
\newblock \href {https://arxiv.org/abs/2602.03782} {{QVLA}: Not all channels
  are equal in vision-language-action model's quantization}.
\newblock \emph{arXiv preprint arXiv:2602.03782}.

\bibitem[{Yan et~al.(2026)Yan, Wan, Ye, Yu, Du, You, and Tsang}]{yan2026hbvla}
Xin Yan, Zhenglin Wan, Feiyang Ye, Xingrui Yu, Hangyu Du, Yang You, and Ivor
  Tsang. 2026.
\newblock \href {https://arxiv.org/abs/2602.13710} {{HBVLA}: Pushing 1-bit
  post-training quantization for vision-language-action models}.
\newblock \emph{arXiv preprint arXiv:2602.13710}.

\bibitem[{Zhang et~al.(2025)Zhang, Liu, Li, Lv, Wang, and
  Bai}]{zhang2025layersensitive}
Feng Zhang, Yanbin Liu, Weihua Li, Jie Lv, Xiaodan Wang, and Quan Bai. 2025.
\newblock \href {https://arxiv.org/abs/2503.06518} {Towards superior
  quantization accuracy: A layer-sensitive approach}.
\newblock \emph{arXiv preprint arXiv:2503.06518}.

\bibitem[{Zhang et~al.(2026)Zhang, Hsieh, Wan, Lin, Wang, Wang, Lei, and
  Zhang}]{zhang2026quantvla}
Jingxuan Zhang, Yunta Hsieh, Zhongwei Wan, Haokun Lin, Xin Wang, Ziqi Wang,
  Yingtie Lei, and Mi~Zhang. 2026.
\newblock \href {https://arxiv.org/abs/2602.20309v4} {{QuantVLA}:
  Scale-calibrated post-training quantization for vision-language-action
  models}.
\newblock \emph{arXiv preprint arXiv:2602.20309}.

\bibitem[{Zheng et~al.(2025)Zheng, Li, Wang, Liu, Kang, Feng, Zheng, Zou, Chen,
  Zeng et~al.}]{zheng2025x}
Jinliang Zheng, Jianxiong Li, Zhihao Wang, Dongxiu Liu, Xirui Kang, Yuchun
  Feng, Yinan Zheng, Jiayin Zou, Yilun Chen, Jia Zeng, et~al. 2025.
\newblock \href {https://arxiv.org/abs/2510.10274} {{X-VLA}: Soft-prompted
  transformer as scalable cross-embodiment vision-language-action model}.
\newblock \emph{arXiv preprint arXiv:2510.10274}.

\bibitem[{Zheng et~al.(2026)Zheng, Cao, Tao, Mao, Zhang, Li, Chen, Zou, Liu,
  Cao, Mei, and Chen}]{zheng2026dyq}
Zihao Zheng, Hangyu Cao, Chibang Tao, Zhihao Mao, Lingyue Zhang, Maoliang Li,
  Jiayu Chen, Hailong Zou, Xuanzhe Liu, Donggang Cao, Hong Mei, and Xiang Chen.
  2026.
\newblock \href {https://arxiv.org/abs/2603.07904v3} {{DyQ-VLA}:
  Temporal-dynamic-aware quantization for embodied vision-language-action
  models}.
\newblock \emph{arXiv preprint arXiv:2603.07904}.

\end{thebibliography}

\clearpage
\appendix

\begin{table*}[t]
\centering
\small
\setlength{\tabcolsep}{3pt}
\begin{tabular}{@{}llrrrl@{}}
\toprule
Model & Group & Layers & Params (M) & Share (\%) & QbD-VLA \\
\midrule
\multirow{8}{*}{$\pi_{0.5}$} & Vision encoder & 162 & 411.1 & 14.6 & W4A6 \\
& Language backbone & 126 & 1981.8 & 70.2 & W4A6 \\
& Action head: attention projections & 72 & 84.9 & 3.0 & W4A4 \\
& Action head: pre-attention modulators & 18 & 56.6 & 2.0 & W4A6 \\
& Action head: post-attention and final modulators & 18 & 56.6 & 2.0 & W4A4 \\
& Action head: MLP layers & 54 & 226.5 & 8.0 & W4A4 \\
& Action head: action and time projections & 5 & 5.3 & 0.2 & W4A4 \\
& \textit{All groups} & 455 & 2822.9 & 100.0 & W4.00A5.74 \\
\midrule
\multirow{7}{*}{OpenVLA-OFT} & Vision encoder & 209 & 728.3 & 9.8 & W4A6 \\
& Vision-language projector & 5 & 88.2 & 1.2 & W4A4 \\
& Language backbone & 224 & 6476.0 & 87.0 & W4A6 \\
& Action head: input projection \texttt{fc1} & 1 & 117.4 & 1.6 & W4A4 \\
& Action head: residual-block layers & 2 & 33.6 & 0.5 & W4A4 \\
& Action head: output projection \texttt{fc2} & 1 & 0.029 & $<$0.001 & BF16 \\
& \textit{All groups} & 442 & 7443.5 & 100.0 & W4.00A5.94 \\
\midrule
\multirow{8}{*}{X-VLA} & Vision encoder & 96 & 335.0 & 42.4 & W4A6 \\
& Language backbone & 72 & 151.0 & 19.1 & W4A4 \\
& Action head: attention QKV projections & 24 & 75.5 & 9.6 & W4A8 \\
& Action head: attention output projections & 24 & 25.2 & 3.2 & W8A8 \\
& Action head: MLP \texttt{fc1} layers & 24 & 100.7 & 12.7 & W4A6 \\
& Action head: MLP \texttt{fc2} layers & 24 & 100.7 & 12.7 & W4A6 \\
& Action head: remaining quantizable layers & 2 & 2.1 & 0.3 & FP32 \\
& \textit{All groups} & 266 & 790.1 & 100.0 & W4.16A5.90 \\
\midrule
\multirow{6}{*}{$\pi_0$} & Vision encoder & 162 & 411.1 & 15.2 & W4A4 \\
& Language backbone & 126 & 1981.8 & 73.2 & W4A4 \\
& Action head: attention projections & 72 & 84.9 & 3.1 & W8A8 \\
& Action head: MLP layers & 54 & 226.5 & 8.4 & W4A8 \\
& Action head: action and time projections & 5 & 3.2 & 0.1 & BF16 \\
& \textit{All groups} & 419 & 2707.6 & 100.0 & W4.14A4.47 \\
\bottomrule
\end{tabular}
\caption{Layer groups used for bit allocation, with the number of quantizable linear layers in each group, their parameter count, their share of the model's quantizable parameters, and the level QbD-VLA selects for the group on LIBERO-Spatial. BF16 and
FP32 mark groups kept at the released precision. The \textit{All groups} rows give the parameter-weighted nominal bit widths used in Table~\ref{tab:libero}.}
\label{tab:groups}
\end{table*}

\section{Layer Groups and Selected Allocations}
\label{app:groups}

Bit allocation operates on groups of linear layers rather than on individual layers.
Table~\ref{tab:groups} lists the groups of each model, together with the level QbD-VLA selects for each group on LIBERO-Spatial.
They follow the component analysis of VLAQuantBench \citep{vlaquantbench}: the vision encoder, the vision-language projector where it exists as a separate module, and the language backbone each form one group, and the action head is split by layer role.
For $\pi_{0.5}$, the action expert's self-attention projections, its pre-attention adaptive-RMSNorm modulators, its post-attention and final modulators, its MLP layers, and the action and time projections form five groups.
$\pi_0$ has the same attention, MLP, and projection groups and no adaptive-RMSNorm modulators.
The action head of OpenVLA-OFT is an MLP whose input projection \texttt{fc1}, two residual-block layers, and output projection \texttt{fc2} form three groups, the last being a single layer of 28{,}672 parameters.
The 24 action-head blocks of X-VLA contribute their QKV projections, attention output projections, MLP \texttt{fc1} layers, and MLP \texttt{fc2} layers, and the two remaining quantizable layers of the head form the last group.
The language-model output head and the convolutional patch embeddings are never quantized.

All layers in a group share one level, and the group definitions remain fixed during allocation.
The nominal cost counts a group at released precision as 16 weight and 16 activation bits, including the X-VLA groups that execute in FP32.
The language backbone holds 70 to 87\% of the quantizable parameters of $\pi_{0.5}$, OpenVLA-OFT, and $\pi_0$, so the nominal weight cost is set almost entirely by its level, whereas the action-head groups hold between less than 0.001\% and 13\% of the parameters and are where the selected allocations differ from uniform settings.
Across the four policies, the vision encoder and language backbone stay at W4A6 or W4A4. 
Every level above W4A6, and every group kept at released precision, is an action-head group.

\section{Experimental Configuration}
\label{app:config}

\begin{sloppypar}
$\pi_{0.5}$ and $\pi_0$ are the LeRobot LIBERO fine-tunes (\nolinkurl{lerobot/pi05_libero_finetuned_v044} and \nolinkurl{lerobot/pi0_libero_finetuned_v044}, BF16), OpenVLA-OFT uses the released per-suite checkpoints
(\nolinkurl{moojink/openvla-7b-oft-}\allowbreak\nolinkurl{finetuned-libero-}\{\nolinkurl{spatial}, \nolinkurl{object}, \nolinkurl{goal}, \nolinkurl{10}\}, BF16) with the allocation selected on the Spatial checkpoint applied to the other three, and
X-VLA uses one released checkpoint for all four LIBERO suites (\nolinkurl{2toINF/X-VLA-Libero}) and one for SIMPLER (\nolinkurl{2toINF/X-VLA-Google-Robot}), both in FP32.
Each model runs in its official evaluation code, wrapped by one benchmark runner that fixes the initial states, episode limits, and success criterion.
Quantization is simulated in PyTorch with round-to-nearest and no calibration data: 4-bit weights use asymmetric per-group quantization with group size 128, 8-bit weights symmetric per-channel quantization, and activations symmetric per-token dynamic
absmax scaling at the input of every quantized linear layer.
\end{sloppypar}

Table~\ref{tab:config} lists the per-model allocation settings.
Levels are ordered FP $\to$ W8A8 $\to$ W4A8 $\to$ W4A6 $\to$ W4A4.
Each batch runs one episode per task, ten per batch, with initial state $20+i$ in batch $i \in \{0,1,2\}$, and a candidate is evaluated for at most three batches.
The full-precision reference is estimated from the same three batches, and the floor is $\tau$ times that estimate, with $\tau = 0.75$ for $\pi_{0.5}$ and $0.90$ for the other models. 
OpenVLA-OFT selects the same allocation under either value.
Acceptance and rejection use Wilson bounds with $z_{\mathrm{acc}} = 1.28$ and $z_{\mathrm{rej}} = 1.96$, final selection uses $z = 1.28$, allocation does not start if the reference estimate is below $s_{\min} = 0.5$.
Evaluation results are cached per configuration and reused when a configuration recurs. 
The $\pi_{0.5}$ run additionally reuses 1{,}070 episodes recorded by an earlier run.
The LIBERO-Goal runs for $\pi_{0.5}$ and OpenVLA-OFT use the same settings, and the SIMPLER runs allocate on the variant-aggregation split with a 120-step horizon, with 24 tasks at $\tau = 0.90$ (3{,}072 recorded episodes) and 40 tasks at $\tau = 0.95$
(6{,}560 cached episodes, 800 of them new in the final invocation).
All reported LIBERO numbers follow the official protocol: the released initial-state files, 10 settling steps after reset, $256 \times 256$ renders rotated to upright ($360 \times 360$ for the LeRobot policies, as in their evaluation code), and success
given by the environment flag.
Each test evaluates initial states 0 to 19 of every task, 200 episodes per suite. Held-out tests use states 30 to 49. The full-horizon re-evaluation of accepted configurations uses states 0 to 9, 100 episodes per configuration. 
The allocation states 20
to 22 are disjoint from all of these.
All simulation runs use seed 0 with initial states fixed by the files, so a cell is reproducible given the policy.
The SIMPLER visual-matching split is evaluated on all 864 episodes of its four task families, pick coke can (300), move near (240), open or close drawer (216, six scenes), and put in drawer (108), with the official episode limits.

\begin{table}[t]
\centering
\small
\setlength{\tabcolsep}{3pt}
\begin{tabular}{@{}lcccc@{}}
\toprule
& $\pi_{0.5}$ & OFT & X-VLA & $\pi_0$ \\
\midrule
Released precision & BF16 & BF16 & FP32 & BF16 \\
Render size & 360 & 256 & 256 & 360 \\
Action mode & delta & delta & absolute & delta \\
$H_{\mathrm{alloc}}$ & 120 & 120 & 300 & 120 \\
$H_{\mathrm{eval}}$, Spatial & 280 & 220 & 800 & 280 \\
$H_{\mathrm{eval}}$, Object & 280 & 280 & 800 & 280 \\
$H_{\mathrm{eval}}$, Goal & 300 & 300 & 800 & 300 \\
$H_{\mathrm{eval}}$, Long & 520 & 520 & 900 & 520 \\
$\tau$ & 0.75 & 0.90 & 0.90 & 0.90 \\
New allocation episodes & 970 & 790 & 880 & 600 \\
Allocation wall-clock (h) & 2.6 & 2.5 & 5.1 & 1.6 \\
\bottomrule
\end{tabular}
\caption{Per-model settings. $H_{\mathrm{alloc}}$ is the episode limit during allocation and $H_{\mathrm{eval}}$ the limit of each model's official evaluation code, used for every reported number. Allocation episodes and wall-clock time are for the
primary LIBERO-Spatial runs.}
\label{tab:config}
\end{table}

ActQuant is evaluated from its released 3.09-bit $\pi_{0.5}$ checkpoint in its own C++ runtime, served through a policy server to the same LIBERO runner with a new action chunk requested every 5 steps, and its full-precision reference is the FP16 export of the same checkpoint in the same runtime. 
HoloQ-VLA is evaluated from its released W4A4 and FP16 $\pi_{0.5}$ checkpoints in the openpi inference service it ships with, after 20 warm-up episodes per cell, with the same replanning and runner.
Both use initial states 0 to 19 and the $\pi_{0.5}$ episode limits of Table~\ref{tab:config}.
On the real robot, X-VLA and $\pi_{0.5}$ are fine-tuned with LoRA on the same 411 demonstrations of a Franka FR3 arm picking a named cube (214 nominal and 197 with perturbed approach paths), and the LoRA weights are merged before the frozen
LIBERO-Spatial allocation is applied.
The client sends absolute end-effector commands at 15\,Hz, X-VLA executes 20 predicted actions per inference and $\pi_{0.5}$ 10, a trial ends after 300 control steps, the first gripper output above 0.5 is latched as the close command, and the
instruction is always ``pick up the \{color\} cube''.
X-VLA runs 30 fixed three-cube layouts per condition. $\pi_{0.5}$ runs six layouts per condition and six per
uniform diagnostic setting.
Success requires the instructed cube to be lifted at least 6\,cm for one second, and a safety controller clips commands to the workspace, limits each step to 6\,cm, and ends a trial as a failure when the intervention counter exceeds 30.
Simulation experiments run on one RTX 4090 and one laptop RTX 5080.

\end{document}